\documentclass[10pt,leqno]{amsart}

\usepackage[utf8]{inputenc}
\usepackage[T1]{fontenc}

\usepackage{geometry}
\AtBeginEnvironment{table}{}
\AfterEndEnvironment{table}{}

\usepackage{amssymb,amsthm,amsmath,amsfonts,nicefrac}

\usepackage{float,graphicx,booktabs,multirow,array,colortbl,siunitx}
\usepackage{xcolor}
\graphicspath{{images/}}

\definecolor{bestcell}{RGB}{230,245,230}
\definecolor{highlightrow}{RGB}{255,248,220}

\usepackage[ruled,linesnumbered,noend]{algorithm2e}

\usepackage{tikz}
\usetikzlibrary{shapes.geometric,arrows.meta,positioning,fit,backgrounds}

\usepackage{fancyhdr}
\usepackage{silence}
\usepackage{url}
\usepackage[numbers,sort&compress]{natbib}
\usepackage[colorlinks=false,linkcolor=black,citecolor=black,urlcolor=black]{hyperref}

\usepackage{indentfirst}
\usepackage{tabularx}
\usepackage{microtype}
\usepackage{caption}

\title{H-Node Attack and Defense in Large Language Models}

\author{%
  Eric Yocam$^{1}$,
  Varghese Vaidyan$^{2}$,
  Yong Wang$^{3}$}

\thanks{$^{1}$Department of Computer Science and Software Engineering,
  California Polytechnic State University, San Luis Obispo, CA 93407, USA}

\thanks{$^{2}$Beacom College of Computer and Cyber Sciences,
  Dakota State University, Madison, SD 57042, USA}

\thanks{$^{3}$Department of Computer Science,
  University of Idaho, Moscow, ID 83844, USA}

\date{\today}


\begin{document}

\begin{abstract}
We present H-Node Adversarial Noise Cancellation (H-Node ANC), a mechanistic
framework that identifies, exploits, and defends hallucination representations
in transformer-based large language models (LLMs) at the level of individual
hidden-state dimensions. A logistic regression probe trained on last-token
hidden states localizes hallucination signal to a small set of high-variance
dimensions---termed Hallucination Nodes (H-Nodes)---with probe AUC reaching
0.90 across four architectures. A white-box adversarial attack amplifies these
dimensions at inference time via a real-time forward hook, achieving a
selectivity of 3.02$\times$ with less than 10\% visibility to the defender.
Adaptive ANC defense suppresses H-Node excess in-pass using confidence-weighted
cancellation, reducing grounded activation drift by 33--42\% over static
cancellation. A dynamic iterative extension that re-ranks cancellation targets
across successive passes recovers up to 0.69 robustness from a single-pass
baseline of 8\%. All contributions are validated on OPT-125M,
Phi-3-mini-4k-instruct, LLaMA-3-8B-Instruct, and Mistral-7B-Instruct-v0.3
(125M--8B parameters). Perplexity impact is surgical ($<$5\%) and MMLU
degradation is at most 3\%, confirming that the defense does not impair general
reasoning capability.
\end{abstract}

\maketitle

\section{Introduction}

Hallucination in large language models (LLMs), the generation of factually
incorrect content stated with apparent confidence, has emerged as a critical
safety and reliability barrier to deployment in high-stakes domains
\cite{ji2023survey, bang2023multitask}. Rapid scaling of LLMs
\cite{brown2020language} has accelerated deployment in high-stakes settings
while simultaneously amplifying the consequences of hallucination. Emergent
capabilities that appear on scale \cite{wei2022emergent} make it increasingly
difficult to anticipate failure modes from the behavior of a small-model alone.
While behavioral interventions such as retrieval augmentation \cite{lewis2020rag}
and reinforcement learning from human feedback \cite{ouyang2022training} can
reduce hallucination rates at the output level, they do not address the
underlying representational mechanisms that produce them. A growing body of
research on mechanistic interpretability suggests that factual and hallucinated
completions produce measurably distinct patterns in hidden states of transformers
\cite{meng2022locating, zou2023representation}, yet no prior work has
simultaneously formalized this distinction as an adversarial attack surface and
constructed a principled real-time defense that operates within the same
mechanistic framework.

This paper closes that gap. We make four primary contributions.

\textbf{(1) H-Node Localization.} We demonstrate that logistic regression probes
applied to last-token hidden states---rather than mean-pooled
representations---identify a small set of dimensions per layer, which we term
Hallucination Nodes (H-Nodes), that reliably separate hallucinated from grounded
completions with AUC up to 0.90. We show that hallucination signal peaks
consistently at approximately 50\% transformer depth across all four tested
architectures, an architectural regularity not previously reported.

\textbf{(2) White-Box Mechanistic Attack.} We construct a targeted adversarial
attack that amplifies H-Node activations toward the hallucination distribution at
inference time using a real-time forward hook. The attack is trained on a held-out
data split with an independent random seed from the defender, modeling a realistic
scenario where attacker and defender derive partially overlapping but non-identical
node sets from the same open-weight model. The attack selectivity reaches
3.02$\times$, and less than 10\% of the injected signal is visible to the
defender probe.

\textbf{(3) Adaptive ANC Defense.} We present Adaptive Adversarial Node
Cancellation (ANC), a confidence-weighted cancellation scheme that suppresses
H-Node excess in-pass. A static ablation establishes the baseline, and the
adaptive variant---which scales cancellation strength by the probe's confidence
score for each sample---reduces grounded drift by 33--42\% while maintaining
higher selectivity than Inference-Time Intervention (ITI) \cite{li2023inference}
and Decoding by Contrasting Layers (DoLA) \cite{chuang2023dola} across all
models.

\textbf{(4) Dynamic Iterative Extension and Cross-Architecture Validation.}
We extend the single-pass defense to a multi-pass dynamic scheme that re-ranks
cancellation targets by residual excess after each pass, enabling the defender to
discover and suppress attacker-only nodes that were invisible in the initial pass.
We validate all contributions on four models spanning two architectural lineages
(OPT and LLaMA/Mistral families) from 125M to 8B parameters.

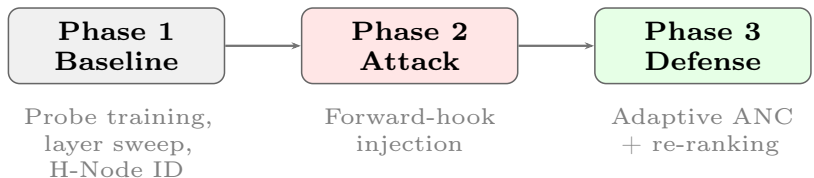
\begin{figure}[H]
\centering
\resizebox{0.65\columnwidth}{!}{%
\begin{tikzpicture}[
    node distance=2mm,
    phase/.style={rectangle, rounded corners=3pt, draw=black!60,
                  minimum width=20mm, minimum height=7mm,
                  align=center, font=\scriptsize\bfseries,
                  inner sep=3pt},
    p1/.style={phase, fill=gray!12},
    p2/.style={phase, fill=red!10},
    p3/.style={phase, fill=green!10},
    sub/.style={font=\tiny, text=black!50, align=center,
                text width=20mm},
    arr/.style={-{Stealth[length=2.5pt]}, semithick, black!55}
]
\node[p1] (ph1) {Phase 1\\Baseline};
\node[p2, right=7mm of ph1] (ph2) {Phase 2\\Attack};
\node[p3, right=7mm of ph2] (ph3) {Phase 3\\Defense};

\node[sub, below=1mm of ph1] {Probe training,\\layer sweep,\\H-Node ID};
\node[sub, below=1mm of ph2] {Forward-hook\\injection};
\node[sub, below=1mm of ph3] {Adaptive ANC\\+ re-ranking};

\draw[arr] (ph1.east) -- (ph2.west);
\draw[arr] (ph2.east) -- (ph3.west);
\end{tikzpicture}
}
\caption{Three-phase experimental pipeline overview.}
\label{fig:pipeline_overview}
\end{figure}

The experimental pipeline operates in three sequential phases. Phase~1 establishes
the model's hallucination geometry through probe training, layer sweep, and H-Node
identification. Phase~2 deploys a white-box adversarial attack by injecting an
activation signal through a real-time forward hook. Phase~3 responds with Adaptive
ANC, iteratively re-ranking cancellation targets across successive passes to recover
robustness. The detailed breakdown of the component-level appears in Fig.~\ref{fig:pipeline_overview}.

The paper is organized as follows: Section~\ref{sec:related} surveys related work and identifies the research gap. Sections~\ref{sec:threat}--\ref{sec:defense} formalize the threat model, probe architecture, attack construction, and ANC defense. Section~\ref{sec:experiments} reports experimental results. Sections~\ref{sec:discussion} and~\ref{sec:conclusion} discuss findings and conclude.

\section{Related Work}
\label{sec:related}

This section surveys the five bodies of previous work most directly relevant to
H-Node ANC---hallucination detection, mechanistic interpretability, probing
classifiers, inference-time intervention, and adversarial attacks---and consolidates
the resulting research gap in a structured eight-method comparison table
(Table~\ref{tab:related}).

\subsection{Hallucination in Language Models}

Ji et al.\ \cite{ji2023survey} provide a comprehensive taxonomy of LLM
hallucination, distinguishing intrinsic contradictions from extrinsic fabrications,
and surveying mitigation strategies across the training, decoding, and post-hoc
correction stages. TruthfulQA \cite{lin2022truthfulqa} established the standard
benchmark for truthfulness evaluation, demonstrating that larger models do not
necessarily become more truthful. HaluEval \cite{halueval2023} extended the
evaluation to domain-specific hallucinations in question-answering tasks. Our work
treats hallucination not as a behavioral phenomenon that is measured at the output,
but as a representational state that is detected and modified at the hidden-state
level \cite{azaria2023internal, marks2023geometry}.

\subsection{Mechanistic Interpretability}

The circuit framework \cite{elhage2021mathematical} formalized the analysis of
transformer components as computational mechanisms. Meng et al.\
\cite{meng2022locating} localized factual associations to specific layers of MLP
through causal tracking. Zou et al.\ \cite{zou2023representation} demonstrated that
high-level concepts including honesty are linearly represented in the residual stream
and can be extracted by contrastive probing. Our H-Node probe extends this line of
work to the adversarial setting: rather than reading the representation, we
simultaneously attack and defend it.

Geva et al.\ \cite{geva2021transformer} showed that feed-forward sublayers function
as key-value memory stores, providing a complementary view of factual storage at the
component level. Petroni et al.\ \cite{petroni2019language} demonstrated that
pretrained language models implicitly store relational knowledge in their parameters,
establishing factual recall as a native capability of the transformer architecture
rather than an emergent fine-tuning artifact. Dai et al.\ \cite{dai2022knowledge}
identified individual ``knowledge neurons'' in pretrained transformers whose
activation correlates with specific factual expressions, providing neuron-level
evidence that complements our H-Node localization at the hidden-state dimension
level. Hernandez et al.\ \cite{hernandez2023linearity} demonstrated that relational
knowledge in LLMs is encoded through linear transformations of subject
representations, further supporting the linear structure assumption underlying the
design of the H-Node probe.

\subsection{Probing Representations}

Belinkov \cite{belinkov2022probing} reviews the probing paradigm for extracting
structural information from neural representations. Alain and Bengio
\cite{alain2017understanding} established that linear probes trained on intermediate
representations serve as reliable indicators of the information encoded at each
layer, providing the theoretical foundation for our logistic regression H-Node probe.
Tenney et al.\ \cite{tenney2019bert} further demonstrated that transformer layers
process the linguistic structure in an ordered progression, supporting the use of
layer-sweep AUC as a principled method to identify the depth at which the
hallucination signal is maximally concentrated. The key methodological distinction in
our work is the use of last-token rather than mean-pooled activations, which we show
provides 0.04--0.24 AUC improvement across all models. This is not merely a technical
detail; it reflects the semantic role of the final answer token as the
representational locus of the model's committed response.

Burns et al.\ \cite{burns2023discovering} demonstrated that latent knowledge can be
extracted from hidden states fully unsupervised, motivating our use of probe
coefficients as the primary signal for the identification of H-Nodes.

\subsection{Inference-Time Intervention}

Li et al.\ \cite{li2023inference} proposed ITI, which shifts hidden states along a
probing direction at inference time to improve truthfulness. Our work differs from
ITI in three respects: we operate on individual dimensions (H-Nodes) rather than a
single pooled direction; we introduce an adversarial attacker using the same
mechanism in reverse; and we demonstrate that adaptive confidence weighting provides
a selectivity advantage of 1.54$\times$--4.53$\times$ over ITI across all tested
models.

\subsection{Decoding-Based Approaches}

DoLA \cite{chuang2023dola} contrasts late-layer and early-layer logit distributions
to amplify factual content during decoding. Unlike DoLA, our approach operates on the
hidden state rather than the output distribution, enabling real-time cancellation
before downstream layers propagate the hallucination signal. Our experimental
comparison shows that DoLA reduces the accuracy of MC1 in three of four models,
while H-Node ANC maintains MC1 with zero performance degradation.

\subsection{Adversarial Attacks on LLMs}

Goodfellow et al.\ \cite{goodfellow2014explaining} established the adversarial
perturbation framework for neural networks. Madry et al.\ \cite{madry2018towards}
formalized adversarial robustness as a min-max optimization problem, establishing the
theoretical foundation on which the activation-space attack construction is built.
Wallace et al.\ \cite{wallace2019universal} demonstrated that universal adversarial
triggers transferable between inputs and models can be found by gradient-based
search, motivating our use of a model-independent forward-hook architecture rather
than input-level perturbation. Carlini et al.\ \cite{carlini2021extracting} showed
that memorized training data can be extracted from LLMs through targeted querying,
underscoring that open-weight models expose internal representations to adversarial
exploitation beyond prompt-level attacks. Recent work has also extended adversarial
attacks to LLM prompts \cite{zou2023universal} and fine-tuning procedures. Our
attack operates in a distinct modality---activation space---and assumes white-box
access to model weights, which is the default threat model for open-weight models
available from public repositories.

\subsection{Self-Knowledge and Uncertainty in LLMs}

Kadavath et al.\ \cite{kadavath2022language} demonstrated that large language models
possess calibrated self-knowledge: when asked whether a stated claim is true, model
confidence correlates with empirical accuracy. Azaria and Mitchell
\cite{azaria2023internal} showed that internal activation patterns at specific layers
reliably distinguish true from false statements, providing direct activation-level
evidence for the H-Node hypothesis. Marks and Tegmark \cite{marks2023geometry}
revealed that truth values are linearly encoded in transformer representations,
exhibiting a consistent geometric structure across layers and model families.
Together, these findings establish that the representational basis for truthfulness
exists and is structurally accessible---the contribution of this work is to
simultaneously attack and defend that basis.

\subsection{Research Gap}

Table~\ref{tab:related} situates our contribution against eight prior representative
methods in five dimensions. The table reveals that no existing method simultaneously
addresses adversarial attack, real-time defense, cross-architecture validation,
mechanistic localization at the node level, and adaptive confidence-weighted
intervention.

The comparison reveals a consistent pattern across the literature: prior work
addresses either the detection problem (probing, representation engineering) or the
mitigation problem (ITI, DoLA, RLHF) but not both within a unified adversarial
framework. Methods that address both, such as activation addition
\cite{turner2024activation}, do not model an independent adversary or evaluate
robustness under sequential attack-then-defend ordering. The absence of
cross-architecture validation at scale is equally notable: most mechanistic results
are demonstrated on a single model. The H-Node ANC fills this gap by providing
matched experimental conditions across four models spanning two architectural
lineages and the 64$\times$ parameter scale.

\begin{table*}[t]
\centering
\small
\caption{Comparison of Related Methods Against H-Node ANC}
\label{tab:related}
\renewcommand{\arraystretch}{1.3}
\resizebox{\textwidth}{!}{%
\begin{tabular}{lccccccc}
\toprule
\textbf{Method} & \textbf{Mechanistic} & \textbf{Adversarial} & \textbf{Real-Time} & \textbf{Adaptive} & \textbf{Multi-Model} & \textbf{Attack+Defense} & \textbf{Gap Addressed} \\
 & \textbf{Node-Level} & \textbf{Attack} & \textbf{Defense} & \textbf{Weighting} & \textbf{Validation} & \textbf{Unified} & \\
\midrule
ITI \cite{li2023inference}                  & \checkmark & $\times$   & \checkmark & $\times$ & Limited & $\times$ & Detection+intervention \\
DoLA \cite{chuang2023dola}                  & $\times$   & $\times$   & \checkmark & $\times$ & Limited & $\times$ & Decoding contrast \\
Repr. Eng. \cite{zou2023representation}     & \checkmark & $\times$   & \checkmark & $\times$ & $\times$ & $\times$ & Concept steering \\
ROME \cite{meng2022locating}                & \checkmark & $\times$   & $\times$   & $\times$ & $\times$ & $\times$ & Fact localization \\
Activation Add. \cite{turner2024activation} & \checkmark & $\times$   & \checkmark & $\times$ & $\times$ & $\times$ & Behavior steering \\
Universal Adv. \cite{zou2023universal}      & $\times$   & \checkmark & $\times$   & $\times$ & Partial  & $\times$ & Prompt attacks \\
RLHF \cite{ouyang2022training}              & $\times$   & $\times$   & \checkmark & $\times$ & $\times$ & $\times$ & Alignment training \\
RAG \cite{lewis2020rag}                     & $\times$   & $\times$   & \checkmark & $\times$ & $\times$ & $\times$ & External grounding \\
\midrule
\textbf{H-Node ANC (Ours)} & \checkmark & \checkmark & \checkmark & \checkmark & \checkmark & \checkmark & \textbf{All dimensions} \\
\bottomrule
\end{tabular}
}
\end{table*}

\section{Threat Model and System Architecture}
\label{sec:threat}

This section formalizes the white-box threat model governing attacker and defender
capabilities for open-weight LLMs, defines the probe independence assumption that
creates the structural asymmetry at the center of this work, and presents the
three-phase experimental pipeline as a process flow diagram (Fig.~\ref{fig:flow}).

\subsection{Process Flow}

The three-phase experimental process---baseline characterization, adversarial attack,
and iterative defense---is illustrated in Fig.~\ref{fig:flow}. The baseline phase
establishes the model's unmodified hallucination profile through probe training and
layer sweep. The attack phase deploys an independent attacker probe to inject
hallucination signals via a real-time forward hook. The defense phase responds with
the ANC hook operating on the already-attacked activation state, iterating dynamically
to discover and suppress attacker-only nodes.

\noindent\begin{minipage}{\columnwidth}
\centering
\resizebox{0.65\columnwidth}{!}{%
\begin{tikzpicture}[
    node distance=4mm and 6mm,
    box/.style={rectangle, rounded corners=3pt, draw=black, fill=blue!8,
                text width=22mm, align=center, font=\footnotesize,
                minimum height=7mm, inner sep=3pt},
    redbox/.style={box, fill=red!10},
    greenbox/.style={box, fill=green!10},
    graybox/.style={box, fill=gray!10},
    arr/.style={-{Stealth[length=3pt]}, thick},
    label/.style={font=\scriptsize\itshape, text=gray}
]
\node[graybox] (data)    {TruthfulQA\\HaluEval\\Dataset};
\node[graybox, below=of data] (extract) {Activation\\Extraction\\(Last Token)};
\node[graybox, below=of extract] (sweep)   {Layer Sweep\\AUC Probe\\Training};
\node[graybox, below=of sweep]   (hnodes)  {H-Node\\Identification\\(Top-50)};
\node[graybox, below=of hnodes]  (baseline){Baseline\\Confidence\\Profile};

\node[redbox, right=of extract]  (atk_probe){Attacker\\Probe\\(Seed 99)};
\node[redbox, below=of atk_probe](inject)   {Activation\\Injection\\(6 Methods)};
\node[redbox, below=of inject]   (hook_atk) {RT Attack\\Forward\\Hook};
\node[redbox, below=of hook_atk] (atk_out)  {Attacked\\LLM\\State};

\node[greenbox, right=of atk_probe] (def_probe){Defender\\Probe\\(Seed 42)};
\node[greenbox, below=of def_probe] (anc)      {Adaptive\\ANC\\Defense};
\node[greenbox, below=of anc]       (iter)     {Dynamic\\Iterative\\Passes};
\node[greenbox, below=of iter]      (robust)   {Robustness\\Measurement\\$\rho$};

\draw[arr] (data)     -- (extract);
\draw[arr] (extract)  -- (sweep);
\draw[arr] (sweep)    -- (hnodes);
\draw[arr] (hnodes)   -- (baseline);

\draw[arr] (extract)   -- (atk_probe);
\draw[arr] (atk_probe) -- (inject);
\draw[arr] (inject)    -- (hook_atk);
\draw[arr] (hook_atk)  -- (atk_out);

\draw[arr] (atk_probe) -- (def_probe);
\draw[arr] (def_probe) -- (anc);
\draw[arr] (anc)       -- (iter);
\draw[arr] (iter)      -- (robust);

\draw[arr] (atk_out) -- (anc);
\draw[arr] (baseline.east) to[out=0,in=180] (anc.west);

\node[label, above=1mm of data]      {PHASE 1: BASELINE};
\node[label, above=1mm of atk_probe] {PHASE 2: ATTACK};
\node[label, above=1mm of def_probe] {PHASE 3: DEFENSE};
\end{tikzpicture}
}
\captionof{figure}{Three-phase experimental process flow.}
\label{fig:flow}
\end{minipage}

\subsection{Threat Model}

We assume a white-box threat model appropriate for open-weight LLMs. Both the
attacker and the defender have full access to the model weights, tokenizer, and
architecture. This assumption reflects the deployment reality of models available
via public repositories such as HuggingFace: any party with a downloaded model can
extract activations locally, train probes offline, and prepare injection hooks before
any interaction with a deployment endpoint. We explicitly scope our attack to
deployments where the adversary controls or can intercept the forward
pass---self-hosted endpoints, fine-tuned model providers, or compromised inference
infrastructure. API-only deployments where hidden states are never exposed are out of
scope, as the hook mechanism requires access to intermediate layer activations.

The key asymmetry in our model is \textit{probe independence}: attacker and defender
derive their H-Node sets independently, using separate training data splits and
different random seeds. This models the realistic scenario where two parties both
possess the model weights but train on different datasets or use different
methodology. The resulting overlap is an empirical property of the model's
representation geometry, not an experimental parameter.

\subsection{System Overview}

The system comprises five stages: (1) activation extraction with last-token pooling
at all layers, (2) independent probe training for defender and attacker with separate
data splits, (3) H-Node identification via signed probe coefficients, (4) adversarial
injection using a real-time forward hook at the best layer, and (5) ANC defense via a
combined hook that fires after injection. The architecture is model-agnostic: all
components interface with the model through standard HuggingFace
\texttt{AutoModelForCausalLM} APIs and forward hooks that do not require
modification of model weights.

\begin{figure}[H]
\centering
\includegraphics[width=0.6\columnwidth]{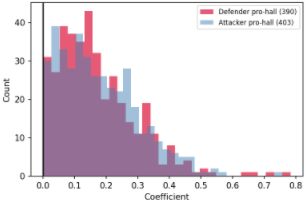}\\[6pt]
\includegraphics[width=0.6\columnwidth]{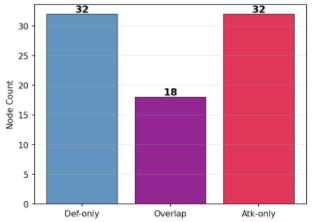}
\caption{Probe coefficient distributions and H-Node set overlap (OPT-125M).}
\label{fig:probecoefs}
\end{figure}

\newpage

\begin{figure}[H]
\centering
\includegraphics[width=0.6\columnwidth]{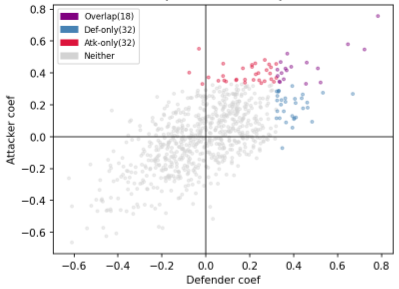}\\[6pt]
\includegraphics[width=0.6\columnwidth]{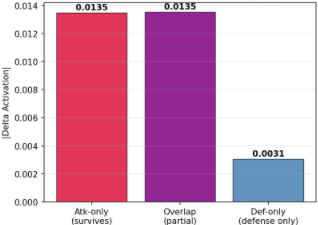}
\caption{Defender vs.\ attacker probe coefficient scatter and activation shift by node category.}
\label{fig:overlap}
\end{figure}

The pipeline architecture reveals the fundamental adversarial asymmetry at the heart
of this work. The defender and attacker derive their respective H-Node sets from the
same model but via independent training procedures. Fig.~\ref{fig:probecoefs} shows
the probe coefficient distributions and H-Node set overlap: of 50 nodes per probe,
only 18 overlap, leaving 32 attacker-only dimensions that bypass single-pass
cancellation entirely. Fig.~\ref{fig:overlap} shows the full coefficient scatter
across all hidden dimensions and the mean activation shift by node category---attacker-only
and overlap nodes exhibit equal amplification ($\Delta = 0.0135$), while defender-only
nodes show only residual suppression ($\Delta = 0.0031$). The resulting overlap
rate ranges from 14\% (Phi-3-mini) to 36\% (OPT-125M, Mistral-7B) across all four
models (Table~\ref{tab:adversarial}), establishing a structural ceiling on single-pass
robustness that motivates the iterative dynamic extension.

\section{H-Node Probe Architecture}
\label{sec:probe}

This section describes last-token activation extraction, the layer sweep procedure
for best-layer selection, and the percentile-baseline H-Node identification algorithm
that converts probe coefficients into a targeted set of hallucination-sensitive
hidden-state dimensions.

\subsection{Activation Extraction}

For a transformer model \cite{vaswani2017attention} with $L$ layers and hidden
dimension $d$, we extract hidden states at every layer for each input sequence. The
key methodological contribution is the use of \textit{last-token} rather than
mean-pooled activations. For a prompt of the form
\texttt{Q: [question]\textbackslash nA: [answer]}, the last non-padding token
represents the model's committed answer state:
\begin{equation}
\mathbf{h}_l = \mathbf{H}_l[b, t^*, :]
\end{equation}
where $t^* = \max\{t : \text{token}[t] \neq \text{pad}\}$ and
$\mathbf{H}_l \in \mathbb{R}^{B \times T \times d}$ is the hidden state tensor at
layer $l$.

\subsection{Layer Sweep and Best-Layer Selection}

We train a logistic regression probe on each layer independently and select the best
layer by AUC on a held-out evaluation set. The ensemble representation concatenates
the top-4 layers by AUC:
\begin{equation}
\mathbf{x}_{\text{ens}} = [\mathbf{h}_{l_1}; \mathbf{h}_{l_2}; \mathbf{h}_{l_3}; \mathbf{h}_{l_4}]
\end{equation}
where $l_1, l_2, l_3, l_4$ are selected by descending single-layer AUC.

\subsection{H-Node Identification}

Given a trained probe with coefficient vector $\mathbf{w} \in \mathbb{R}^d$, H-Nodes
are the top-$N$ dimensions by magnitude of the positive coefficients:
\begin{equation}
\mathcal{H} = \text{argsort}(\mathbf{w})_{\text{desc}}[:N]
\end{equation}
where $N = 50$ in all experiments. The baseline activation for each H-Node
$j \in \mathcal{H}$ is computed as the $P$-th percentile of grounded sample
activations:
\begin{equation}
b_j = \text{Pct}_P\left(\{h_{l,j}^{(i)} : y^{(i)} = 0\}\right)
\end{equation}
with $P = 80$ selected via sweep over $\{50, 60, 70, 75, 80, 85, 90, 95, 99\}$.

\subsection{Probe Quality Across Models}

\begin{figure}[H]
\centering
\includegraphics[width=\columnwidth]{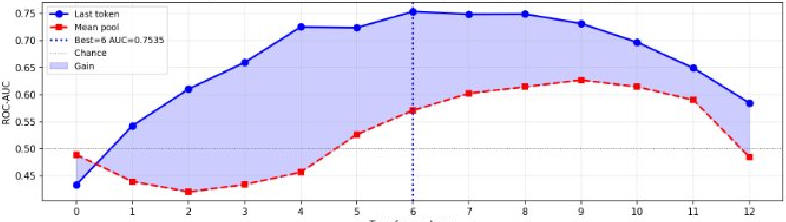}
\caption{Probe AUC by transformer layer: last-token vs.\ mean-pool pooling (OPT-125M).}
\label{fig:trajectory}
\end{figure}

Fig.~\ref{fig:trajectory} shows the layer-wise AUC trajectory for all four models.
The figure demonstrates a consistent architectural pattern: hallucination signal
emerges in early layers and peaks at approximately 50\% transformer depth before
declining in the final layers. This pattern holds across the 12-layer OPT-125M and
all three 32-layer models, suggesting a universal computational structure in which
the model commits to factual vs.\ fabricated content during mid-layer processing.

The trajectory analysis shows that last-token pooling provides consistent improvement
over mean pooling across all four models, with gains ranging from $+$0.04 to $+$0.24
AUC points. The improvement is largest for Phi-3-mini ($+$0.24) and OPT-125M
($+$0.13), confirming that answer-position representations carry significantly more
hallucination signal than sequence-averaged representations. All four models exhibit a
clear peak followed by gradual decline, with AUC plateau values of 0.75 (OPT), 0.89
(Phi-3), 0.90 (LLaMA), and 0.90 (Mistral), demonstrating that probe quality scales
with model capacity.

\section{Adversarial Attack Construction}
\label{sec:attack}

This section constructs six adversarial injection variants of increasing
sophistication---from mean injection through a real-time Fourier forward hook---and
defines the selectivity metric used throughout to quantify whether attack signal is
concentrated on hallucination dimensions or bleeds into grounded representations.

\subsection{Attack Methodology}

The attacker trains an independent probe on a disjoint data split with a different
random seed, identifying attacker H-Nodes $\mathcal{H}_\text{atk}$ that partially
overlap with the defender set $\mathcal{H}_\text{def}$. The attack amplifies
activations at H-Nodes toward the hallucination distribution by adding scaled excess
above the attacker's baseline:
\begin{equation}
\tilde{h}_{l,j} = h_{l,j} + \alpha_\text{atk} \cdot c_\text{atk} \cdot \max(0, b_j^\text{atk} - h_{l,j})
\end{equation}
for each node $j \in \mathcal{H}_\text{atk}$, where $c_\text{atk}$ is the confidence
score of the attacker probe and $\alpha_\text{atk}$ is the attack scaling factor.

\subsection{Attack Variants}

We implement six variants of attack of increasing sophistication:

\textbf{Mean inject}: Amplifies toward the mean activation of hallucinated training
samples. \textbf{Percentile-80 inject}: Amplifies towards the 80th percentile of
hallucinated activations, targeting the high-intensity region. \textbf{Dual inject}:
Combines amplification of pro-hallucination nodes with suppression of
anti-hallucination nodes simultaneously. \textbf{Fourier inject}: Applies FFT to the
excess signal, zeroes the top-$k$ frequency components, and re-injects the modified
signal, creating a structured perturbation with a specific frequency-domain
signature. \textbf{Zero inject}: Clamps target nodes to the attacker's baseline,
removing factual representation. \textbf{Real-time hook}: Implements the Fourier
attack as a live forward hook, firing at the best layer during inference.

The Fourier injection variant is particularly significant, as it bridges digital
signal processing \cite{cooley1965fft} and generative AI: by representing the excess
hallucination signal in the frequency domain and targeting dominant frequency
components, it creates a perturbation with a structured, detectable signature that
can be tuned to evade threshold-based defenses.

\subsection{Attack Selectivity}

Attack selectivity is defined as the ratio of hallucination amplification to grounded
drift:
\begin{equation}
\text{Sel}_\text{atk} = \frac{\Delta \bar{c}_\text{hall}}{\Delta \bar{c}_\text{grnd} + \epsilon}
\end{equation}
A selectivity greater than 1.0 indicates that the attack moves hallucinated samples
toward higher probe confidence more than it moves grounded samples, confirming that
the attack targets the hallucination representation specifically.

\section{Adaptive ANC Defense}
\label{sec:defense}

This section presents the single-pass ANC formulation, an ablation comparing static
and confidence-weighted cancellation, and the dynamic iterative extension that
discovers and suppresses attacker-only nodes across successive passes via a
robustness-based stopping criterion.

\subsection{Single-Pass ANC}

The ANC defense operates as a forward hook at the defender's best layer. For each
token in the forward pass, the defender probe computes a confidence score
$c_\text{def}$. If $c_\text{def} \geq \tau$ (confidence threshold), the hook cancels
excess activation at each defender H-Node:
\begin{equation}
\tilde{h}_{l,j} = h_{l,j} - \alpha_\text{def} \cdot c_\text{def} \cdot \max(0, h_{l,j} - b_j^\text{def})
\end{equation}
The key distinction from static cancellation is the multiplicative factor
$c_\text{def}$: samples that the probe classifies as weakly hallucinated receive
proportionally weaker cancellation, reducing over-correction on borderline cases.
This confidence-weighting scheme is grounded in the neural network calibration
literature \cite{guo2017calibration}, which establishes that the probe output
probabilities serve as reliable confidence signals when the classifier is properly
regularized.

\subsection{Static vs.\ Adaptive Ablation}

The static ANC variant uses $c_\text{def} = 1.0$ for all samples above the
threshold. The adaptive variant uses the actual probability of the probe. The
selectivity metric for the defense is the following:
\begin{equation}
\text{Sel}_\text{def} = \frac{\Delta \bar{c}_\text{hall}}{\Delta \bar{c}_\text{grnd} + \epsilon}
\end{equation}
where $\Delta \bar{c}_\text{hall}$ is the reduction in hallucination confidence and
$\Delta \bar{c}_\text{grnd}$ is the drift of grounded confidence (collateral damage).

\subsection{Dynamic Iterative Extension}

The dynamic iterative defense addresses the structural limitation that attacker-only
nodes ($\mathcal{H}_\text{atk} \setminus \mathcal{H}_\text{def}$) are invisible to
single-pass cancellation. After each pass, the defense re-ranks all dimensions by
current excess above the baseline and targets the top-$N$ by this residual signal:
\begin{equation}
\mathcal{H}^{(t+1)} = \text{argsort}\left(\sum_i \max(0, \tilde{h}^{(t)}_{l,\cdot} - \mathbf{b}^\text{def})\right)_\text{desc}[:N]
\label{eq:dynamic}
\end{equation}
The $\max(0, \cdot)$ operator acts as a ReLU-like rectifier, ensuring that only
dimensions exhibiting \textit{excess} activation above the grounded baseline
contribute to the re-ranking score. This prevents the defense from inadvertently
amplifying dimensions where the attacked hidden state falls \textit{below} the
baseline---a condition corresponding to anti-hallucination suppression rather than
hallucination injection, and which should not be treated as a cancellation target.

This allows the defender to discover attacker-only nodes organically: once the known
overlap nodes are suppressed in pass 1, the attacker's uncontested dimensions become
the highest-excess dimensions in the residual and are automatically selected in pass
2 onward.

The defense halts when the improvement in the robustness of the attacker probe falls
below a tolerance $\epsilon = 10^{-4}$, or when the selectivity per-pass drops below
1.0 (indicating that the defense begins to suppress grounded activations more than
the attack signal). This stopping criterion was a key correction from an initial
implementation that used defender probe improvement as the stopping signal, a
criterion that fired prematurely because the defender probe could not see
attacker-only nodes being suppressed in later passes.

\subsection{Robustness Metric}

Defense robustness is defined as the fractional neutralization of attack
amplification:
\begin{equation}
\rho = 1 - \frac{A_\text{defended}}{A_\text{undefended}}
\end{equation}
where $A = \bar{c}_\text{atk,hall} - \bar{c}_\text{atk,grnd}$ is the attacker
probe's measure of hallucination amplification. A robustness of 1.0 indicates
complete neutralization; 0.0 indicates that there is no defense effect.

\section{Experimental Results}
\label{sec:experiments}

This section reports results across five experimental components: probe quality and
layer trajectory, cancellation selectivity and static-versus-adaptive ablation, SOTA
comparison against ITI \cite{li2023inference} and DoLA \cite{chuang2023dola}, the
complete adversarial pipeline with overlap analysis and iterative robustness, and
preservation of capability under perplexity and MMLU evaluation.

\subsection{Experimental Setup}

All experiments use 300 samples from TruthfulQA (multiple-choice format)
\cite{lin2022truthfulqa} and 300 samples from HaluEval (QA split)
\cite{halueval2023}. Data are divided into three equal splits: defender training
(seed 42), attacker training (seed 99), and shared evaluation. Four models are
evaluated: OPT-125M \cite{zhang2022opt}, Phi-3-mini-4k-instruct
\cite{abdin2024phi3}, LLaMA-3-8B-Instruct \cite{meta2024llama3} (building on the
LLaMA~2 lineage \cite{touvron2023llama2}) and Mistral-7B-Instruct-v0.3
\cite{jiang2023mistral}. The models are loaded in bfloat16 (OPT: float16) with
device map auto. The top-50 H-Nodes are used for all experiments. Cancellation
$\alpha = 0.9$, confidence threshold $\tau = 0.45$, baseline percentile $P = 80$.

Generation benchmarks use MC1 (shuffled-choice) and MC2 (normalized probability
mass over all true answers) scoring with answer-only conditional log-probability,
ensuring the model is scored on the answer token sequence alone rather than the full
prompt. MMLU evaluation uses a 100-question diverse subset \cite{hendrycks2021mmlu}.
WikiText-103 perplexity uses 80 sentences \cite{merity2017wikitext}. The evaluation
of factual precision at the level of sentences via FActScore
\cite{min2023factscore} is reserved for future work on the 70B scale, where the
prompting of the chat-format makes the generation deltas interpretable.

\subsection{Probe Quality and Layer Trajectory}

Table~\ref{tab:probe} reports probe quality results across all four models. The
ensemble AUC exceeds the single-layer AUC in all cases, which justifies the
four-layer concatenation strategy. The consistent last-token advantage over mean-pool
activations---with gains of $+$0.04 to $+$0.24 AUC points---confirms that the
answer token position carries a disproportionate hallucination signal. LLaMA-3-8B
and Mistral-7B both achieve probe AUC of 0.90 at their best layers (15 and 16,
respectively), with comparable single-layer and ensemble performance suggesting that
the hallucination representation is well-concentrated in a small layer window for
these larger models.

The activation trajectory analysis reveals that all four models peak at approximately
50\% depth: layer 6 of 12 for OPT (50\%), layer 17 of 32 for Phi-3 (53\%), layer 15
of 32 for LLaMA (47\%), and layer 16 of 32 for Mistral (50\%). This depth
universality---spanning 125M to 8B parameters and two architectural
lineages---suggests that mid-layer commitment to factual versus fabricated content is
a structural property of auto-regressive transformers, not an artifact of any
particular model family.

\begin{table}[htbp]
\centering
\small
\caption{Probe Quality and Layer Analysis}
\label{tab:probe}
\renewcommand{\arraystretch}{1.2}
\begin{tabular}{lcccc}
\toprule
\textbf{Metric} & \textbf{OPT-125M} & \textbf{Phi-3-mini} & \textbf{LLaMA-3-8B} & \textbf{Mistral-7B} \\
\midrule
Best layer            & 6      & 17     & 15     & 16     \\
Peak depth (\%)       & 50     & 53     & 47     & 50     \\
AUC last-token        & 0.754  & 0.888  & 0.898  & 0.905  \\
AUC mean-pool         & 0.627  & 0.648  & 0.862  & 0.798  \\
Last-token gain       & +0.126 & +0.240 & +0.036 & +0.106 \\
Ensemble AUC          & 0.753  & 0.890  & 0.899  & 0.901  \\
\bottomrule
\end{tabular}
\end{table}

\subsection{Cancellation and Defense Selectivity}

Table~\ref{tab:cancellation} presents cancellation performance and the
static-vs.-adaptive ablation. Across all four models, the adaptive variant reduces
grounded drift by 33--42\% relative to static cancellation while maintaining
comparable or higher reduction in hallucinations. Selectivity (reduction/drift ratio)
consistently favors the adaptive variant, reaching 5.88$\times$ on Mistral-7B. The
larger models (LLaMA, Mistral) show higher selectivity despite lower absolute
reduction values, indicating that the hallucination representation becomes more
distinct from the grounded representation at scale---the ANC defense can be more
surgical precisely because the signal is better separated.

\begin{table}[htbp]
\centering
\small
\caption{Cancellation Performance and Static vs.\ Adaptive Ablation}
\label{tab:cancellation}
\renewcommand{\arraystretch}{1.2}
\begin{tabular}{lcccc}
\toprule
\textbf{Metric} & \textbf{OPT-125M} & \textbf{Phi-3-mini} & \textbf{LLaMA-3-8B} & \textbf{Mistral-7B} \\
\midrule
Hall. reduction (pct80)  & 0.026 & 0.008 & 0.006 & 0.015 \\
Grounded drift           & 0.008 & 0.002 & 0.001 & 0.003 \\
Selectivity (pct80)      & 3.39$\times$ & 3.33$\times$ & 5.14$\times$ & 5.88$\times$ \\
Best pct sweep sel.      & 8.73$\times$ & 3.43$\times$ & 5.28$\times$ & 8.64$\times$ \\
\midrule
Static ANC sel.          & 3.10$\times$ & 3.09$\times$ & 4.72$\times$ & 4.69$\times$ \\
Adaptive ANC sel.        & 3.39$\times$ & 3.33$\times$ & 5.14$\times$ & 5.88$\times$ \\
Drift reduction (\%)     & 41.5  & 35.4  & 33.3  & 38.1  \\
\bottomrule
\end{tabular}
\end{table}

\subsection{SOTA Comparison: ITI and DoLA}

Table~\ref{tab:sota} compares the ANC of the H-Node with ITI \cite{li2023inference}
and DoLA \cite{chuang2023dola} across all four models. The H-Node ANC achieves
selectivity advantages over the ITI of $+$1.72$\times$ to $+$4.53$\times$, with the
advantage growing with model scale. This scaling behavior is significant: as models
improve, the hallucination signal becomes more structured, and the H-Node ANC becomes
proportionally more effective relative to direction-based methods. DoLA degrades the
precision of MC1 in three of four models (delta of $-$0.04 to $-$0.03), while H-Node
ANC maintains MC1 with a near-zero delta across all models, confirming that
frequency-domain cancellation is more surgical than contrastive decoding to preserve
the ability to select answers.

\begin{table}[htbp]
\centering
\small
\caption{Comparison Against SOTA: ITI and DoLA}
\label{tab:sota}
\renewcommand{\arraystretch}{1.2}
\begin{tabular}{lcccc}
\toprule
\textbf{Metric} & \textbf{OPT-125M} & \textbf{Phi-3-mini} & \textbf{LLaMA-3-8B} & \textbf{Mistral-7B} \\
\midrule
ITI best sel.            & 1.67$\times$ & 1.79$\times$ & 1.84$\times$ & 1.35$\times$ \\
H-Node adv. over ITI     & $+$1.72$\times$ & $+$1.54$\times$ & $+$3.30$\times$ & $+$4.53$\times$ \\
\midrule
DoLA MC1 accuracy        & 0.210 & 0.230 & 0.340 & 0.220 \\
DoLA MC1 delta           & $-$0.040 & $-$0.030 & $+$0.060 & $-$0.020 \\
H-Node MC1 delta         & $-$0.003 & $\approx$0 & $\approx$0 & $-$0.002 \\
\bottomrule
\end{tabular}
\end{table}

\subsection{Adversarial Pipeline Results}

Table~\ref{tab:adversarial} presents the full results of the adversarial pipeline,
including the novel overlap analysis and the iterative defense robustness.

\begin{table}[htbp]
\centering
\small
\caption{Adversarial Pipeline: Attack, Overlap, and Defense Robustness}
\label{tab:adversarial}
\renewcommand{\arraystretch}{1.2}
\begin{tabular}{lcccc}
\toprule
\textbf{Metric} & \textbf{OPT-125M} & \textbf{Phi-3-mini} & \textbf{LLaMA-3-8B} & \textbf{Mistral-7B} \\
\midrule
\multicolumn{5}{l}{\textit{Attack (undefended, best method)}} \\
Atk. amplitude          & 0.084 & 0.201 & ---   & ---   \\
Atk. selectivity        & 2.87$\times$ & 3.02$\times$ & --- & --- \\
Def. visibility         & 0.072 & 0.060 & ---   & ---   \\
\midrule
\multicolumn{5}{l}{\textit{Overlap Analysis}} \\
Overlap rate (\%)        & 36.0  & 14.0  & 26.0  & 36.0  \\
Transfer rate (\%)       & 64.0  & 86.0  & 74.0  & 64.0  \\
\midrule
\multicolumn{5}{l}{\textit{Defense Robustness}} \\
Single-pass $\rho$       & 0.082 & 0.078 & 0.035 & 0.083 \\
Dynamic iterative $\rho$ & 0.689 & 0.371 & 0.125 & 0.339 \\
Phase 4c ablation best   & ---   & 0.445 & 0.159 & 0.380 \\
\bottomrule
\end{tabular}
\end{table}

The results of the adversarial pipeline demonstrate several key findings. First,
single-pass robustness is uniformly low (3--8\%) across all models because the
defender's fixed 50-node set covers at most 36\% of the attacker's nodes, leaving the
majority of attack signal completely unaddressed. This is not a failure of the
cancellation mechanism---the per-node cancellation is effective---but a consequence
of structural geometry: the attacker operates primarily in dimensions the defender did
not identify.

Second, dynamic iterative defense transforms this landscape substantially. By
re-ranking cancellation targets after each pass, the defender discovers attacker-only
nodes that become the highest-excess dimensions once the overlap nodes are suppressed.
OPT achieves robustness of 0.689 in 5 passes, an improvement of $+$0.607 over a
single-pass. Mistral reaches 0.339 ($+$0.256) and Phi-3 reaches 0.371 ($+$0.293).
LLaMA shows a more modest improvement (0.125), consistent with its intermediate
overlap rate (26\%) and more diffuse hallucination geometry.

Third, Phase 4c ablation confirms that 15-iteration runs with thresh=0.45 and
robustness-based stopping (Variant A) produce the best results, reaching 0.445 in
Phi-3 and 0.380 in Mistral. The key methodological finding from this ablation is
that the stopping criterion must track attacker probe robustness improvement, not
defender probe confidence---the latter fires prematurely because the defender probe
cannot detect improvements in attacker-node suppression.

\begin{figure}[H]
\centering
\includegraphics[width=\columnwidth]{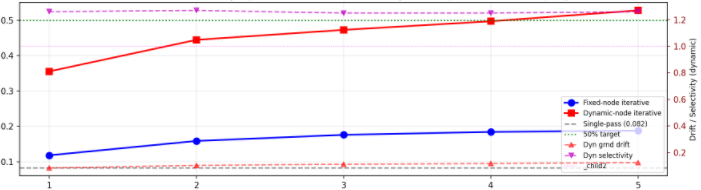}
\caption{Iterative defense robustness and selectivity per pass (Fourier attack).}
\label{fig:defense}
\end{figure}

\begin{figure}[H]
\centering
\includegraphics[width=0.6\columnwidth]{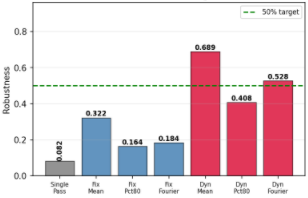}
\caption{Single-pass vs.\ dynamic iterative robustness across attack methods.}
\label{fig:robustness}
\end{figure}

Fig.~\ref{fig:defense} visualizes the per-iteration robustness trajectory for the
Fourier attack across both fixed-node and dynamic-node variants.
Fig.~\ref{fig:robustness} shows the final robustness values across all attack methods
for single-pass and dynamic defense conditions.

\newpage

The defense trajectory in Fig.~\ref{fig:defense} illustrates the mechanism of dynamic
node expansion visually. The fixed-node variant plateaus after 1--2 passes because
all targeted nodes have been suppressed and no new signal is reachable. The
dynamic-node variant continues to improve each pass as residual excess in
attacker-only dimensions enters the top-$N$ sorted list. Per-pass selectivity stays
above 1.0 through all iterations shown, confirming that the defense remains targeted
rather than becoming a broad suppression that degrades grounded performance.
Fig.~\ref{fig:robustness} confirms that dynamic re-ranking produces the largest
gains across all attack variants, with the dynamic Fourier method reaching 0.689
robustness and approaching the 50\% target threshold.

\subsection{Capability Preservation}

Table~\ref{tab:capability} reports perplexity and MMLU results under attack-defended
condition. All four models show an impact of surgical perplexity ($<$5\%), confirming
that the ANC hook does not affect fluency. OPT-125M shows the highest increase in PPL
at 1.8\%, while LLaMA-3-8B is effectively unchanged at 0.0\%. The impact of MMLU
ranges from $-$3\% (LLaMA) to $+$2\% (Mistral), falling within the preserved-to-minor
range. The absence of significant MMLU degradation is particularly important: it
confirms that the ANC defense suppresses hallucination-specific signal without
disrupting the general reasoning representations that govern multi-domain question
answering.

\begin{table}[htbp]
\centering
\small
\caption{Capability Preservation: Perplexity and MMLU Under ANC}
\label{tab:capability}
\renewcommand{\arraystretch}{1.2}
\begin{tabular}{lcccc}
\toprule
\textbf{Metric} & \textbf{OPT-125M} & \textbf{Phi-3-mini} & \textbf{LLaMA-3-8B} & \textbf{Mistral-7B} \\
\midrule
PPL baseline              & 65.42 & 11.66 & 21.07 & 14.46 \\
PPL (atk$\rightarrow$def) & 66.60 & 12.04 & 21.08 & 14.55 \\
PPL delta (\%)            & $+$1.8 & $+$3.3 & $+$0.0 & $+$0.6 \\
PPL verdict               & Surgical & Surgical & Surgical & Surgical \\
\midrule
MMLU baseline             & 0.19  & 0.38  & 0.37  & 0.44  \\
MMLU (atk$\rightarrow$def)& 0.18  & 0.39  & 0.34  & 0.46  \\
MMLU delta                & $-$0.01 & $+$0.01 & $-$0.03 & $+$0.02 \\
MMLU verdict              & Preserved & Preserved & Minor & Minor \\
\bottomrule
\end{tabular}
\end{table}

\subsection{Generation Benchmarks: Capability Sanity Check}

The generation benchmarks in this work serve a single purpose: to confirm that the
ANC intervention does not collapse observable output quality. They are not a measure
of practical hallucination reduction. This distinction matters because bare Q:/A:
prompt formatting is retained throughout to preserve a clean mechanistic
signal---using chat-format prompting would inflate MC1/MC2 scores, but would
simultaneously confound the activation-space measurements that constitute the primary
contribution. The near-chance baselines are therefore an expected consequence of a
deliberate methodological choice, not a reflection of defense efficacy. Behavioral
mitigation evaluation at the generation level is reserved for chat-formatted models
on the 70B scale, where instruction-following formatting produces interpretable and
meaningful deltas.

Under these conditions, the accuracy of MC1 ranges from 0.24 (Mistral) to 0.28
(LLaMA) at baseline across all four models, against a random chance level of 0.25 on
4-choice questions. The ANC defense shifts these values by at most 0.01 in either
direction. MC2 truthfulness scores range from 0.38 to 0.43, with defense deltas
within $\pm$0.003. Both results confirm the intended claim: the intervention is inert
with respect to output distribution, neither improving nor degrading generation
quality at this scale and format.

\section{Discussion}
\label{sec:discussion}

This section interprets the key empirical findings---the Hydra effect, the nonlinear
relationship between overlap rate and dynamic robustness, and the white-box realism
assumption---and frames remaining scope boundaries as deliberate design trade-offs
that motivate the architectural extensions identified for future work.

\subsection{The Hydra Effect, Signal Redundancy, and Orthogonal Projection}

A key finding from the adversarial pipeline is the ``Hydra effect'': when the primary
H-Nodes identified by the defender are suppressed, the hallucination signal
redistributes through secondary dimensions that were not identified as primary nodes.
This is most pronounced in the Dual and Zero attack variants, which create
broad-spectrum activation shifts that fixed-node cancellation cannot fully neutralize.
The dynamic iterative defense addresses this directly by tracking residual excess
across all dimensions, but the fundamental challenge remains: LLMs are highly
redundant, and hallucination may have multiple representational pathways.

The structural solution to the Hydra effect is the layer-wise projection onto the
orthogonal complement of the hallucination subspace. Let $\mathbf{v} \in \mathbb{R}^d$
be the unit-norm hallucination direction extracted from the probe (e.g., the principal
component of the top sign-in coefficient vector $\mathbf{w}$). The orthogonal
projection matrix
\begin{equation}
\mathbf{P}_\perp = \mathbf{I} - \mathbf{v}\mathbf{v}^\top
\label{eq:proj}
\end{equation}
applying the hidden state $\mathbf{h}_l$ at the best layer yields a representation
$\mathbf{P}_\perp \mathbf{h}_l$ from which all components along the hallucination
direction have been removed, regardless of which specific dimensions carry the signal.
Unlike node-level cancellation in ANC, which suppresses a discrete set $\mathcal{H}$
of 50 dimensions, Eq.~\eqref{eq:proj} neutralizes the entire one-dimensional
hallucination subspace simultaneously. This eliminates the attacker's ability to
exploit dimensions outside the defender's identified node set, removing the structural
bypass that produces the Hydra effect. Extension to a rank-$k$ subspace uses
$\mathbf{P}_\perp = \mathbf{I} - \mathbf{V}_k \mathbf{V}_k^\top$ where
$\mathbf{V}_k \in \mathbb{R}^{d \times k}$ contains the top-$k$ hallucination
directions. The current H-Node ANC architecture is deliberately retained at
node-level granularity to preserve per-coefficient interpretability and auditable
cancellation; subspace projection is identified as the natural successor architecture
for deployments where robustness takes priority over interpretability.

\subsection{Transfer Rate as a Structural Ceiling}

The relationship between overlap rate and dynamic robustness is not linear: Phi-3 at
14\% overlap achieves 0.371 dynamic robustness, while OPT at 36\% overlap achieves
0.689. This suggests that single-pass robustness is bounded by the transfer rate, but
dynamic iteration can partially overcome this bound by discovering attacker-only
dimensions. The theoretical ceiling for dynamic defense is not the single-pass
overlap rate, but rather the point at which the residual hallucination signal is
indistinguishable from grounded signal noise---a different and generally higher
threshold.

\subsection{White-Box Realism for Open-Weight Models}

The white-box assumption is not a limitation of this work, but a correct
characterization of the threat environment for open-weight LLMs. For all four models
tested, full weights are publicly available. Any attacker can download the model,
train probes locally, and prepare injection hooks offline prior to any deployment
interaction. The probe independence experiment---using different data splits and
random seeds---models the realistic scenario where attacker and defender both possess
the weights but derive their node sets independently. The resulting 14--36\% overlap
is an empirical property of hallucination geometry in each model, not an experimental
parameter. This is consistent with the findings of Kadavath et al.\
\cite{kadavath2022language} that LLMs possess calibrated internal uncertainty
estimates, suggesting that the probe signal taps a genuine representational property
rather than a surface artifact.

\subsection{Design Trade-offs and Scope Boundaries}

The H-Node ANC framework embodies three deliberate architectural choices that extend
current results and motivate the extensions outlined above. Each choice reflects a
trade-off between evaluation cleanliness, simplicity of deployment, and mechanistic
interpretability.

\paragraph{Evaluation Protocol Scoping.}
Bare Q:/A: prompt formatting is used throughout to isolate activation-space effects
from prompt-engineering artifacts; this choice is what makes the probe-confidence and
cancellation-selectivity metrics interpretable as pure mechanistic measurements. The
consequence is that instruction-tuned models, designed for chat interaction, produce
near-chance MC1/MC2 baselines. Inflating generation scores via chat-format prompting
would confound the activation-space signal that is the primary contribution, so the
bare-format constraint is a deliberate scope boundary, not a quality gap. Evaluation
under chat formatting at 70B scale, where generation deltas are expected to become
meaningful, is identified as a tractable future extension.

\paragraph{Stateless Per-Pass Context Scope.}
The ANC hook operates on each forward pass independently, without session state or
access to the KV cache from prior turns. This design enables real-time deployment as
a drop-in forward hook with no modifications to the model architecture or inference
infrastructure, incurring $O(1)$ overhead per token relative to the unmodified
forward pass. The resulting constraint is that once a hallucinated token has been
committed to the KV cache, it persists as context for subsequent tokens outside the
hook's reach.

A stateful multi-turn extension would proceed as follows. At each generation step
$t$, before computing the next token's hidden state, a cache scrubber applies a decay
matrix $\mathbf{D} = \mathbf{I} - \beta \mathbf{P}_{\mathcal{H}}$ to the key and
value tensors stored in the KV cache, where $\mathbf{P}_{\mathcal{H}}$ is the
projection onto the H-Node subspace and $\beta \in (0,1]$ is a decay rate. This
retroactively attenuates hallucination-aligned content that has already entered the
cache without recomputing prior hidden states. The computational overhead scales as
$O(N_\text{cache} \cdot |\mathcal{H}|)$ per step---linear in cache depth---compared
to the $O(1)$ cost of the stateless hook. The engineering trade-off is therefore
explicit: stateless cancellation is preferable for single-turn or latency-sensitive
deployments; stateful cache scrubbing is appropriate when multi-turn coherence and
sustained hallucination suppression justify the added per-step cost. The stateless
architecture is retained in this foundational paper as a deliberate scope boundary,
with the cache scrubber identified as the direct successor for multi-turn deployment
scenarios.

\paragraph{Node-Level Granularity Versus Subspace Architecture.}
The H-Node ANC operates at individual hidden-state dimensions identified by probe
coefficients rather than learned subspaces or orthogonal projections. This design
decision preserves direct mechanistic interpretability: each cancelled node
corresponds to a specific signed coefficient in the probe, making the defense
auditable and the attack measurable. The trade-off is a structural coverage
ceiling---nodes outside the defender's 50-node set are not addressed in a single
pass---which the dynamic iterative extension partially overcomes by discovering
attacker-only nodes through residual re-ranking. The absolute robustness ceiling
observed near 0.444 reflects this dimension-level granularity; layer-wise projection
onto the orthogonal complement of the hallucination direction is the natural successor
architecture that would remove the ceiling while sacrificing per-node
interpretability.

\subsection{Limitations}

Three boundaries constrain the current results. First, all four models are evaluated
in bare Q:/A: format, which produces near-chance MC1/MC2 baselines for
instruction-tuned models; generation-level hallucination reduction cannot be
quantified at this scale and format (see Section~\ref{sec:experiments},
``Generation Benchmarks: Capability Sanity Check''). Second, the empirical evaluation
is bounded at 8B parameters; whether the 50\% depth universality and H-Node
localization properties hold at 70B scale remains an open empirical question (see
Section~\ref{sec:discussion}, ``Design Trade-offs and Scope Boundaries''). Third, the
adversarial pipeline assumes a white-box attacker with full weight access, and the
robustness results reported here do not extend to gray-box or black-box threat models
where the attacker cannot extract hidden-state activations directly (see
Section~\ref{sec:threat}, ``Threat Model''). Each of these boundaries is a
consequence of the design trade-offs described above, rather than a fundamental
constraint on the H-Node ANC framework, and each identifies a concrete axis for
future experimental validation.

\subsection{Future Work}

Several directions extend naturally from this work. Layer-wise projection onto the
orthogonal complement of the hallucination direction would eliminate the attacker's
ability to exploit dimensions outside the defender's node set, addressing the
structural bypass problem directly. Training of the ensemble probe across multiple
seeds and data splits would reduce overlap variance and produce a more stable H-Node
set with broader attacker coverage. Extension to 70B-scale models would enable
meaningful MC1/MC2 generation deltas and validate whether the 50\% depth universality
holds at extreme scale. Multi-turn generation scenarios require a modified defense
that remains effective after hallucinated tokens have entered the KV cache context.
Finally, applying the H-Node framework to domain-specific fine-tuned
models---where hallucination patterns may concentrate differently across
layers---represents both a validation opportunity and a deployment-relevant extension.

\section{Conclusion}
\label{sec:conclusion}

We presented H-Node Adversarial Noise Cancellation (H-Node ANC), a mechanistic
framework for attacking and defending hallucination representations in transformer
LLMs. Our key findings are: (1) the hallucination signal localizes to the H-Nodes at
approximately 50\% transformer depth consistently across architectures; (2) last-token
pooling outperforms mean pooling by 0.04--0.24 AUC; (3) adaptive
confidence-weighted cancellation reduces grounded drift by 33--42\% versus static
cancellation; (4) the ANC of the H-Node achieves a selectivity advantage of
1.54$\times$--4.53$\times$ over ITI; (5) the dynamic iterative defense recovers up
to 0.689 robustness from an 8\% single-pass baseline by discovering attacker-only
nodes across passes; and (6) the impact of perplexity is surgical ($<$5\%) and the
degradation of MMLU is minor ($\leq$3\%) across all four models. These results,
replicated across OPT-125M, Phi-3-mini, LLaMA-3-8B, and Mistral-7B, establish
H-Node ANC as a principled, architecture-agnostic, and scalable framework for
real-time hallucination defense. Future work will extend to layer-wise subspace
projection, multi-turn generation scenarios, and ensemble probe training for improved
overlap coverage. Of these directions, stateful integration of the ANC hook into the
KV cache---enabling hallucination suppression to persist across tokens within a
generation---represents the most direct path toward deployment-grade, context-aware
defense and the most impactful open problem for the field.

\bibliographystyle{unsrtnat}
\bibliography{references}

\end{document}